\documentclass[11pt,a4paper]{new_template}

\usepackage{algorithm}
\usepackage{kantlipsum, lipsum}
\usepackage{dsfont}
\usepackage{dm-colors}

\usepackage{algorithmic}
\usepackage{threeparttable}
\usepackage{pifont}% http://ctan.org/pkg/pifont
\newcommand{\cmark}{\ding{51}}%
\newcommand{\xmark}{\ding{55}}%
\usepackage{multirow} 
\usepackage[authoryear, sort&compress, round]{natbib}

\author[1]{Zhichao Chen}
\author[1]{Leilei Ding}
\author[1]{Zhixuan Chu}
\author[1]{Yucheng Qi}
\author[1]{Jianmin Huang}
\author[1]{Hao Wang}

\affil[1]{Ant Group, Hangzhou, China}

    \correspondingauthor{Jianmin Huang; Hao Wang}

\title{
      Monotonic Neural Ordinary Differential Equation: Time-series Forecasting for Cumulative Data
      }
    
\begin{abstract}
Time-Series Forecasting based on Cumulative Data (TSFCD) is a crucial problem in decision-making across various industrial scenarios. However, existing time-series forecasting methods often overlook two important characteristics of cumulative data, namely monotonicity and irregularity, which limit their practical applicability. To address this limitation, we propose a principled approach called Monotonic neural Ordinary Differential Equation (MODE) within the framework of neural ordinary differential equations. By leveraging MODE, we are able to effectively capture and represent the monotonicity and irregularity in practical cumulative data. Through extensive experiments conducted in a bonus allocation scenario, we demonstrate that MODE outperforms state-of-the-art methods, showcasing its ability to handle both monotonicity and irregularity in cumulative data and delivering superior forecasting performance.
\end{abstract}


\begin{document}   

% \received{2 June 2023}
% % \received[revised]{4 August 2009}
% \received[accepted]{4 August 2023}

% This command processes the author and affiliation and title
\maketitle
    
\section{Introduction}
\indent Time-Series Forecasting based on Cumulative Data (TSFCD) is of utmost importance in decision-making across various industrial domains. One prominent example is in e-commerce applications, where precise estimation of the cumulative bonus allocated headcount is crucial for optimizing the allocation strategy and enhancing user engagement.

Intuitively, time-series forecasting models~\citep{gluonts_jmlr} can be applied to solve the TSFCD problem. However, these methods overlook two important properties of cumulative data, which may significantly impact their performance in practical applications:
\begin{enumerate}

\item{\textbf{Monotonicity:} Cumulative data always exhibits a monotonically increasing pattern over time. 
This property introduces non-stationary and large variances in the data, making the model training process extremely challenging.
}
    
\item{\textbf{Irregularity:} Inevitable errors, such as those caused by logging systems or sensor systems, result in some recorded data being missing or represented as "not a number" (NAN). 
    % Consequently, the training data becomes irregularly sampled, while the models are typically designed to make inferences from regularly sampled data.
}

\end{enumerate}
    % Failure to address these properties limits the effectiveness of existing time-series forecasting models in practical TSFCD scenarios.
% \indent The former issue will result in large variance, 
\indent To mitigate the issue of large variance caused by monotonicity, existing approaches often resort to forecasting the growth rate~\citep{cramer1962stochastic} instead of the actual cumulative value. Nevertheless, it fails to ensure the monotonicity of the predicted values during the inference stage, leaving ample room for further performance enhancement. Regarding the irregularity issue, one common approach is to incorporate the time difference information into the network. However, since the errors are unknown in advance, the time difference information remains uncertain and can vary significantly, making it challenging for models to capture effective information.
    
\indent To tackle the aforementioned challenges, we begin by analyzing the monotonic increasing property of the predicted values. Through this analysis, we discover that ensuring the monotonicity of the predicted values is more straightforward when modeling the growth rate positively rather than focusing on the exact values. Drawing inspiration from this observation, we transform the time-series forecasting problem into the initial value problem (IVP) of ordinary differential equations (ODEs). We then design numerical solutions to handle irregularities and mitigate the need for explicit time difference scaling. Finally, we propose our Monotonic Ordinary neural Differential Equation (MODE) model within the framework of neural ordinary differential equations (NODE).

    \indent The contributions of this work can be summarized as follows:
    \begin{enumerate}
    \item{We introduce the TSFCD task and reframe it as ODEs IVP, which provides a novel perspective for addressing TSFCD.}
    
    \item{We propose the MODE model, which effectively captures the monotonicity and irregularity of the data. 
    % By incorporating the MODE model within the framework of NODE, we leverage the expressive power of neural networks to accurately forecast cumulative data.
    }
    
    \item{We conduct extensive experiments in simulation, offline, and online environments on a bonus allocation scenario to validate the effectiveness of our MODE model. The results demonstrate significant improvements in forecasting performance compared to state-of-the-art methods.}
    \end{enumerate}
    
    % Overall, our contributions include the formulation of the TSFCD task as an ODE IVP, the development of the MODE model, and the comprehensive evaluation of its performance through various experiments.

    \section{Related Works}\label{sec:related}
    
    \indent Time-series forecasting models have undergone extensive research over the past decades. In order to achieve accurate predictions, numerous neural architectures have been developed to extract informative and predictive features from historical data. These architectures include recurrent neural networks (RNNs) \citep{chung2014empirical} and temporal convolutional networks (TCNs) \citep{bai2018empirical}.
    % each with their own strengths and limitations in terms of modeling capacity and computational efficiency \citep{vaswani2017attention}. 
    In recent years, self-attention networks \citep{vaswani2017attention, wen2022transformers} have emerged as a dominant approach in time series forecasting, following the groundbreaking performance of the LogTrans model \citep{TCNTransformer}. Self-attention networks leverage the power of attention mechanisms to capture the relationships between different time steps in a sequence. This has led to significant advancements in the field, with several extensions and variations proposed, such as Informer \citep{InformerAAAI2021}, FedFormer \citep{zhou2022fedformer}, and AutoFormer \citep{wu2021autoformer}. Each extension brings unique insights and trade-offs, aiming to enhance the applicability and customization of self-attention networks for time series forecasting tasks.\\
    \indent The application of existing methods to TSFCD is hindered by several factors. Firstly, these methods struggle to ensure the monotonic increasing property of their predictions. Secondly, they predominantly assume regularly sampled data, which may not be suitable for practical scenarios where data can be irregularly sampled. A common approach to address the large variance caused by the monotonicity property is to predict the growth rate and obtain the actual values by cumulatively summing the predicted rates. However, even with this approach, the monotonic increasing property of the predictions is not guaranteed, leaving room for performance improvement. Regarding the issue of irregular sampling, previous works have attempted to incorporate time difference information into the models. For instance, the GRU-D model \citep{che2018recurrent} embeds time difference information into the network. 
    % Similarly, models like the continuous kernel convolution network \citep{romero2021ckconv} and flex convolution network \citep{romero2022flexconv} consider time difference information during the model initialization stage. 
    In contrast, time-series models with auto-regressive structures, such as GRU and LSTM, can be viewed as differential equations. Building on this concept, NODE-based models \citep{kidger2020neural} treat time-series forecasting as an ODE IVP \citep{LODEIrregularity}. These models train themselves using the adjoint sensitivity method \citep{chen2018neural}. In this context, the timestamps serve as inputs to the ODE numerical solution algorithms, and the neural network can choose not to explicitly embed the timestamps.\\
    % The timestamp will serve as the parameter in the ODE numerical solution framework rather than the neural network input.
    \indent Based on previous research, we have identified several challenges in addressing the TSFCD problem. In particular, ensuring the monotonically increasing property of the predicted values and effectively handling irregularity in the training data during the training stage are still unresolved challenges. To tackle these issues, we propose our MODE model within the framework of NODE. In the following section, we will provide a comprehensive introduction to the relevant concepts of NODE, which will help to better understand the MODE model and its approach to addressing these challenges.

\section{Preliminaries}\label{Preliminaries}
In this section, we briefly review the IVP of ODEs and numerical solutions to this problem.
Denote the observation value as $y$, which can be modeled by an ODE $f(y(t))$ (we assume the ODE is time-invariant for simplicity). Given the initial value $y(t_0)$ at time $t_0$. The IVP tends to estimate the value at time $t_f$ according to the following equation:
\begin{equation}\label{ODEExpression}
y(t_f) = y(t_0) + \int_{t_0}^{t_f}{f(y(t))\mathrm{d}t}  \approx y(t_0) +  f(y(t_0)) \times (t_f - t_0). 
\end{equation}
Correspondingly, the numerical methods for ODEs attempt to find numerical approximations of the integral term in \eqref{ODEExpression}. Intuitively, as the approximation shows, differential value ($\mathrm{d}t$) can be approximated by difference value ($\Delta t$), and integral operator ($\int$) is then substituted by summation operator ($\sum$) accordingly. 
On this basis, various numerical methods \citep{butcher2016numerical} like Euler method, Adams-Moulton method, Runge-Kutta method and their variants are proposed.
% As a representative approach, the Euler method estimates the integral term as follow:
% % In the Euler method, the integral value between interval $t_0$ and $t_1$ can be estimated by the following equation:
% \begin{equation}\label{EulerMethodExpression}
% \int_{t_0}^{t_1}{f(y(t))\mathrm{d}t}  \approx  f(y(t_0)) \times (t_1 - t_0), %\times 
% \end{equation}
% where the $t_1-t_0$ ($\Delta t$) can be referred to step size in the context of ODEs numerical methods. 
% On this basis, we can estimate the value at time $t_f$ by applying \eqref{EulerMethodExpression} recursively. 
    % as per the line 2-4 in Algorithm \ref{algoEulerMethod}.
    % \begin{algorithm}[htbp]
    %     \caption{Algorithm of Euler Method}
    % \label{algoEulerMethod}
    %     \textbf{Input}: ODE $f_{\theta}(y)$; start point $t_0$; end point $t_f$; step size $\Delta t$; initial value $y(t_0)$.\\
    %     \textbf{Parameter}: ODE function parameter $\theta$\\
    %     \textbf{Output}: Predicted value $y(t_f)$ at time $t_f$.\\
    %     \vspace{-0.3cm}
    %     \begin{algorithmic}[1] %[1] enables line numbers
    %         \STATE Calculate the repeat time $h\leftarrow (t_f - t_0)/{\Delta t}$.
    %         \WHILE{$i < h$}
    %         \STATE $\tau \xleftarrow[]{} t_0+i\times\Delta t$.\\
    %         $y(\tau) \xleftarrow[]{} y({\tau}) + f_{\theta}(y({\tau-\Delta t}))\times \Delta t$
    %         \ENDWHILE
    %         \STATE \textbf{return} $y(t_f)$
    %     \end{algorithmic}
    % \end{algorithm}
    
\indent Note that, the discretization in the time domain might have taken a huge number of steps, and the computation graph may be too big to hold in memory for back-propagation. To address this issue, the adjoint sensitivity method \citep{optimalProcess} is introduced into the NODE model training stage~\citep{chen2018neural}. Define the loss function $\mathcal{L}$ of the NODE model as follow:
\begin{equation}\label{LossfunctionExpression}
\mathcal{L} = \sum_{t=t_0}^{t_N}{ \Vert \tilde{y}(t) - y(t) \Vert^2},
\end{equation}
where $\tilde{y}(t)$ and $y(t)$ are predicted values (obtained by \eqref{ODEExpression}) and real values within time interval $[t_0, t_N]$, respectively.
% where the predicted value $\tilde{y}(t)$ is obtained by \eqref{ODEExpression}.
The adjoint $a(t_i)$ at time $t_i \in [t_0, t_N]$, can be defined as follow:
\begin{equation}
a(t_i) = \frac{\partial{\mathcal{L}}}{\partial{y(t_i)}},
\end{equation}
which obeys the following differential equation~\citep{optimalProcess}:
\begin{equation}
\frac{\mathrm{d} a(t_i)}{\mathrm{d} t} =- {\frac{\partial{f_{\theta}(y(t_i))}}{\partial{y(t_i)}}}^\top a(t_i).
\end{equation}
On this basis, the gradient with respect to model parameter can be obtained as follow:
    \begin{equation}\label{adjointEquation}
    \frac{\partial{\mathcal{L}}}{\partial{\theta}} = -\int_{t_i}^{t_0}{  {\frac{\partial{f_{\theta}(y(t))}}{\partial{\theta}}}^\top a(t) \mathrm{d}t}.
    \end{equation}
Eq. \eqref{adjointEquation} is a new ODE system that does not require us to preserve intermediate value from the forward pass. Based on Eq. \eqref{adjointEquation}, we can augment the original differential equation as follow:
    \begin{equation}
    \begin{aligned}
          \hat{y}(t_0) & = \hat{y}(t_i) + \int_{t_i}^{t_0}{(\frac{\mathrm{d}\hat{y}(t)}{\mathrm{d}t})\mathrm{d}t} 
          =  [y(t_i), \frac{\partial{\mathcal{L}}}{\partial{y(t_i)}}, \mathbf{0}_{\theta}] \\
         & + \int_{t_i}^{t_0}{[f_{\theta}(y(t)), -\frac{\partial{f}_{\theta}}{\partial{y(t)}}^\top a(t), -\frac{\partial{f}_{\theta}}{\partial{\theta}}^\top a(t)]\mathrm{d}t}\\
         &=  [y(t_0), \frac{\partial{\mathcal{L}}}{\partial{y(t_0)}}, \frac{\partial{\mathcal{L}}}{\partial{\theta}}] .
        \end{aligned}
        \label{agumentedFinalState}
    \end{equation}
  Eq. \eqref{agumentedFinalState} suggests that the gradient with respect to the model parameter $\theta$ can be efficiently computed by solving the augmented dynamical system. This eliminates the need to store intermediate values during the numerical solving process in the NODE. 

    \section{Proposed Approach}\label{MethodologySection}
    \subsection{Problem Statement}
    \indent 
    Our objective is to develop a forecasting model that effectively addresses TSFCD problem. The model should be capable of predicting future horizon values ${\tilde{y}(t), \tilde{y}(t+1), ..., \tilde{y}(t+H-1)}$, given a history of known recorded values ${y(t), y(t-1), ..., y(t-W+1)}$, where the recorded values may be irregularly sampled. In our notation, we represent the cumulative data as $y$, the time as $t$, and the model parameters as $\theta$. The designed model should adhere to the following principles:
    
    \begin{enumerate}
    \item{The model predicted values should monotonically increase as time increases.}
    \item{The model should handle the irregularity problem in its structure naturally.}
    \end{enumerate}
    
    \subsection{Growth Rate Forecasting Strategy }\label{MonotonicityIncreasingSection}
    \indent To promise the monotonic increasing property of the predicted values, we first propose the mathematical expression of the monotonic increasing property as follow:
    \begin{equation}\label{MonotonicityExpression}
    y(t_2) - y(t_1) \ge 0, t_2 \ge t_1 \iff \frac{\mathrm{d} y(t)}{\mathrm{d}t} \ge 0.
    \end{equation}
    \indent By observing \eqref{MonotonicityExpression}, we can conclude that it is hard to ensure the neural network satisfies the constraint on the left-hand side. 
    In contrast, the right-hand side of \eqref{MonotonicityExpression} can be easily realized by the non-negative activated function like ReLU function. Therefore, we apply the non-negative activated function ReLU to the output of the neural network:
    \begin{equation}\label{PredictIntegralTerm}
    \frac{\mathrm{d} y(t)}{\mathrm{d}t} = f_{\theta}(y(t)) = \textrm{ReLU}(\textrm{MLP}_{\theta}(y(t))), 
    \end{equation}
    where the MLP stands for the multi-layer perceptron and the subscript $\theta$ is the parameter of the MLP. \\
    \indent Note that, the physical meaning of the first-order derivative is the growth rate. Thus, we convert our modeling object to the growth rate rather than the exact value.
    % it is better for us to model the growth rate rather than the exact value.
    \subsection{ODE Numerical Solving Problem}\label{IrregularityModelingStrategy}
    \indent Based on the previous subsection, we found that modeling the growth rate of cumulative data will be easier to promise the monotonic increasing property of the predicted values.
    % than modeling the cumulative data itself .
    % it would be better to model the growth rate of cumulative data rather than the cumulative data itself to guarantee the monotonic increasing property of the predicted value.
    On this basis, we can obtain the predicted value $\tilde{y}(t_f)$ at time $t_f$ according to the following equation when the initial value $y(t_0)$ at time $t_0$ is given:
    \begin{equation}
    \tilde{y}(t_f) = y(t_0) + \int_{t_0}^{t_f}{  \textrm{ReLU}(\textrm{MLP}_{\theta}(y(t_0))) \mathrm{d}t }.
    \label{ODEIntegrationEquation}
    \end{equation}
    Based on \eqref{ODEIntegrationEquation}, we reformulate the time-series forecasting problem into an ODE IVP, which can be solved by the numerical methods introduced in Section \ref{Preliminaries}.\\
    \indent Since we have converted the time-series forecasting problem into the ODE IVP, intuitively, we can train our MODE model within the NODE framework in Section \ref{Preliminaries}. The numerical methods of the ODEs can handle the irregularly sampled data in the training stage naturally. Specifically, we can drop the NAN during the training stage explicitly to avoid obtaining the gradient on the NAN data. In summary, we address the irregularity issue in the TSFCD problem.
\subsection{Model Overall Architecture}
    
\begin{figure}[htbp]
\centering
\includegraphics[width=0.5\textwidth]{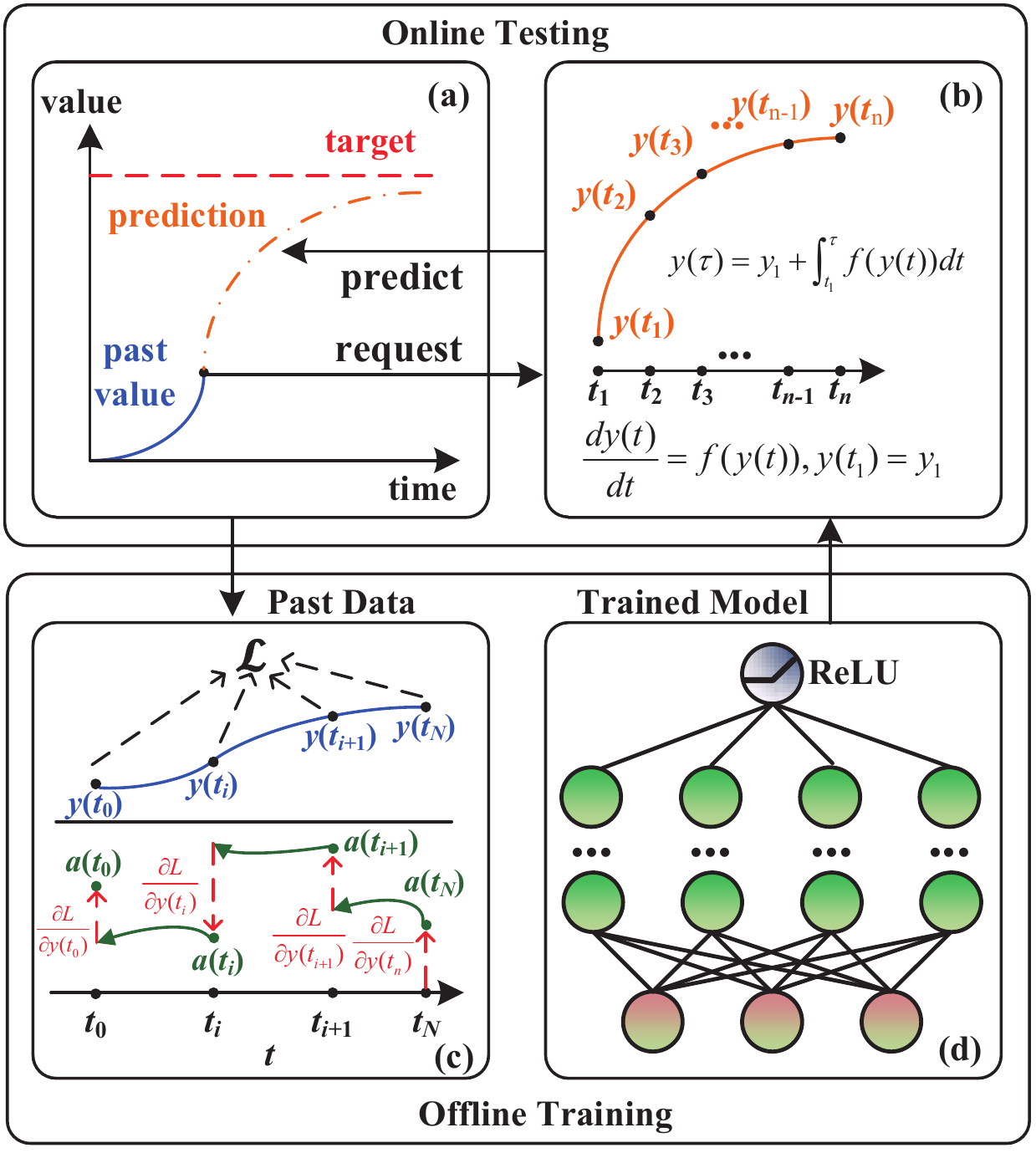}
\caption{The overall illustration of MODE. (a) The model serving environment;% where the blue line indicates the past-known value, the orange line indicates the predicted value, and the red line indicates the target value set from the dwonstream decision task; 
(b) The MODE online inference principle (IVP of ODE); (c) The MODE offline training principle (adjoint sensitivity method); (d) The MODE structure.}

\label{modelStructure}
\end{figure}
Figure \ref{modelStructure} provides an overview of our proposed MODE model, which consists of two main components: online testing and offline training. In Figure \ref{modelStructure} (a), we depict the model's online serving environment, where the model predicts future values (represented by the orange line) based on past values (represented by the blue line). Downstream decisions are made based on the discrepancy between the target value (represented by the red line) and the predicted values. In Figure \ref{modelStructure} (b), we propose that the predicted values are obtained by solving the MODE using numerical methods such as Euler's method. 
For the offline training phase, Figure \ref{modelStructure} (c) illustrates the application of the adjoint sensitivity method for model training. The model's gradient is obtained by solving the backward augmented differential equation as outlined in Eq. \eqref{agumentedFinalState}. Finally, Figure \ref{modelStructure} (d) presents the model structure of the MODE, which is constructed using a Multi-Layer Perceptron (MLP) based on Eq. \eqref{PredictIntegralTerm}.

\section{Experiments}\label{ExperimentSection}
In this subsection, we will conduct experiments to evaluate the performance of MODE and answer the following research questions:
    \\\textbf{Baseline Comparison:}  Does MODE outperform conventional forecasting models (with/without difference value modeling strategy) on cumulative data? 
    % \\\textbf{RQ2:} Does MODE address the monotonic increasing problem of the predicted values better than the difference value modeling strategy?
    % outperform conventional forecasting models when they are designed to model the difference value (growth rate)? % Does MODE outperform conventional forecasting approaches with cumulative sum ($\mathrm{cumsum}$) trick?
    \\\textbf{Ablation Study:} Does the consideration of the monotonic increasing property improve the forecast performance on cumulative data?
    \\\textbf{Irregularity Sampled:} Is MODE robust to irregularly sampled data? 
    % \\\textbf{RQ4 (Sensitivity Analysis):} Does MODE perform well under the changing learning rate and downsampling rate?
    \\\textbf{Online Performance:} Does MODE perform well in the online environment in terms of accuracy and inference time? Besides, what's the MODE model's business value?
    % \\ \textbf{RQ6 (Online Performance): } What's the MODE model business value compared to baseline models?

\subsection{Experimental Settings}
\label{experimentalSettingsPart}
\subsubsection{Background introduction}
\indent To showcase the efficacy of our proposed method, we have chosen a bonus allocation activity from our e-commerce application as an illustrative example. This activity usually occurs two weeks prior to an important traditional Chinese festival. Throughout this activity, users earn virtual cards as they complete tasks within the e-commerce application. The user who manages to collect all specified types of virtual cards is granted a bonus on that Chinese festival. Hence, precise prediction of the daily cumulative headcount of users who acquire all specified of virtual virtual cards plays a vital role in adjusting the strategy for allocating virtual cards.
\subsubsection{Dataset description}
\indent To substantiate the efficacy of our proposed model, we conduct rigorous experiments across three diverse datasets: a synthetic dataset, an offline dataset amassed during the 2021's activity, and an online dataset compiled during the 2022's activity. (\emph{Stringent data protection protocols are in place to minimize the likelihood of unauthorized data duplication or leakage. All datasets have been meticulously cleansed to eliminate personally identifiable information and are safeguarded through robust encryption methods. These datasets are solely intended for research purposes and do not reflect any actual business scenarios.}) The synthetic dataset is generated as follow:
\begin{equation}
    % \begin{aligned}
    y(t)=\zeta\times\exp{(15)} 
    \times\begin{cases}
    1.2\times 10^{-5} \times t,t\in[0,420)\\
    6.0\times10^{-5}(t-420),t\in[420,1440)
    \end{cases},
    % & \zeta\sim Cate(1, 2, 3, ...,10)
    % \end{aligned},
    \label{simulationDataset}
\end{equation}
where $\zeta$ is a random integer belonging to interval $[1, 10]$, $\zeta$ and exponential factor $\exp(15)$ control the scale of every day's total collected headcount.  
% More details about the dataset are .
\subsubsection{Baseline Models}
We consider two groups of baseline models. (1), Autoregressive models: temporal convolution network (TCN)~\citep{bai2018empirical}, Seq2Seq~\citep{sutskever2014sequence}, and Prophet~\citep{taylor2018forecasting}; (2), Non-autoregressive models: Autoformer~\citep{wu2021autoformer} (SOTA approach prior to January, 2022).
\subsubsection{Evaluation protocol}
\indent As per the activity organizers' requirements, the cumulatively collected headcount is reset to 0 at exactly 00:00 hours every day. The organizers will request predictions from the forecasting model every minute. For each time request, they will provide the past recorded value, which includes the data from the previous day along with the recorded data prior to the request time, to the models. Therefore, we can collect a total amount of $60\times24=1440$ points for every day. \\
\indent To assess the accuracy of the forecasts, we adopt the modified mean absolute percentage error ($\textrm{MAPE}$) as the evaluation metric, and the $\textrm{MAPE}$ is calculated for each time request within a future horizon of length $\textrm{H}$ on the testing day, as specified by organizers:
\begin{equation}
      \text{MAPE} = \frac{1}{\textrm{H}}\sum\limits_{t = 1}^{\textrm{H}} {{{\left | \frac {{y(t)} - {{\tilde y}(t)}}{{y(t)}}\right |}}},
      \label{MAPE}
\end{equation}
where $y(t)$ and ${\tilde y}(t)$ are real value and predicted value, respectively. The smaller MAPE indicates better model performance.
We follow the evaluation protocol specified by the activity organizers for all datasets: The model performance is assessed during three intervals: Interval 1 ($[7:00, 14:00)$), Interval 2 ($[14:00, 20:00)$), and Interval 3 ($[20:00, 24:00)$). For evaluation purposes, we choose 2-hour (2H) and 4-hour (4H) horizons as specified by the organizers. However, due to limited data availability in Interval 3, the 2H horizon MAPE is calculated during the time period $[20:00, 22:00]$ merely, and the 4H horizon MAPE is not calculated.\\
\indent Furthermore, in addition to the online evaluation, we utilize the exceed collection rate (ECR) to assess the business value of the models. The ECR is defined by following equation:
\begin{equation}\label{ECRExpression}
  \textrm{ECR} = (\frac{\textrm{Total~Collected~Headcount}}{\textrm{Expected~Collected~Headcount}} - 1) \times 100 \%,
\end{equation}
where "expected collected headcount" is determined by the organizers based on the allocated bonus budget. When the "total collected headcount" exceeds the "expected collected headcount", it indicates that the actual user headcount attracted by the activity is higher than expected. This phenomenon is advantageous for the application, as it indicates a higher level of user engagement. Therefore, a higher ECR value indicates a better performance of the model in the business aspect. To ensure fairness, the same virtual card allocation algorithm is employed for all models.

\subsubsection{Training Protocols}

According to the activity, the production environment will initiate a query request every minute, and the model should provide predictions until the end of the serving day, all within one minute. 
For all datasets, we utilize the day prior to the serving day as the validation dataset, and two days prior to the serving day as the training dataset. On the serving day, we will initialize a total of 1,439 requests, which corresponds to $60\times24-1$.

\indent By  trial and error on the simulation dataset and offline dataset over quantities of time, we set the batch size for the TCN model, Seq2Seq model, and Autoformer model as 64, and the down-sampling rate is set as 20. The batch size for the MODE model is set as 10, which is also the down-sampling rate. The learning rate for the TCN, Seq2Seq, Autoformer, and MODE are set as 0.005, 0.005, and 0.001, 0.0001, respectively. All models are optimized by Adam \citep{kingma2014adam}. 

\indent For the TCN model, we set its layer number as 3, and the channel sizes are [2, 4, 8]. For the Seq2Seq model, we set its layer number as two, and the hidden dimension and LSTM module dimension as 4 and 8, respectively. For the Prophet model, we used the default hyper-parameters provided by the package as initial value, and optimize them automatically by its inner optimization framework. For the Autoformer model, we set its encoder layer and decoder layer as 3 and 2, respectively. Besides, we set the feed-forward dimension and hidden dimension as 8 and 4, respectively. For the MODE model, we set the layer for the MLP as 2, and the hidden units are 2 and 4 for the first and second layers. The nonlinear function is set as $\textrm{ReLU}$. Besides, we model the differential equation in a time-invariant way. In other words, we will not input the time information into the neural network in the MODE.

\indent Note that, since the NODE turns to predict the growth rate. \emph{We also conducted the growth rate forecasting experiment named $\mathrm{cumsum}$ experiment for the above-mentioned models for fairness.} Specifically, we add the \textrm{ReLU} function after the model predicted growth rate. After that, we add the $\mathrm{cumsum}$ operator to the growth rate to obtain the exact time-series predicted value. All hyper-parameters for the baseline models in the $\mathrm{cumsum}$ experiments are the same as aforementioned. To avoid misunderstanding, we will mark the baseline models with $\mathrm{cumsum}$ operator explictly.

  \subsection{Accuracy for Synthetic \& Offline Dataset}
    
    \begin{table}[htbp]
        \centering
        \setlength\tabcolsep{1.8pt}
        \caption{Performance comparison over synthetic dataset.}
   
        \label{NewSimulationData}

    \begin{tabular}{llclllll}
    \hline
    Interval                    & H            & $\mathrm{cumsum}$   & TCN                    & Seq2Seq                & Prophet                                 & Autoformer               & MODE                                                     \\ \hline
    \multirow{4}{*}{Interval 1} & \multirow{2}{*}{2} & \xmark    & 1.86$*$ & 5.48$^*$ & 1.06$*$                  & 1.00$*$ & \multirow{2}{*}{\textbf{0.22}} \\
                                &                    & \cmark & 1.00$*$ & 1.00$*$ & 0.27$*$                  & 1.00$*$ &                                                          \\ \cline{2-8} 
                                & \multirow{2}{*}{4} & \xmark    & 1.66          & 4.55         & 1.16                           & 1.00          & \multirow{2}{*}{0.44}                           \\
                                &                    & \cmark & 1.00          & 1.00$*$ & \textbf{0.27} & 1.00          &                                                          \\ \hline
    \multirow{4}{*}{Interval 2} & \multirow{2}{*}{2} & \xmark    & 1.07$*$ & 2.52$*$ & 0.49$*$                  & 1.00$*$ & \multirow{2}{*}{\textbf{0.09}} \\
                                &                    & \cmark & 1.00$*$ & 1.00$*$ & 0.13$*$                  & 1.00$*$ &                                                          \\ \cline{2-8} 
                                & \multirow{2}{*}{4} & \xmark    & 1.02          & 2.27         & 0.53                           & 1.00         & \multirow{2}{*}{0.17}                           \\
                                &                    & \cmark & 1.00          & 1.00$*$ & \textbf{0.14} & 1.00          &                                                          \\ \hline
    \multirow{2}{*}{Interval 3} & \multirow{2}{*}{2} & \xmark    & 0.90$*$ & 1.94$*$ & 0.36$*$                  & 1.00$*$ & \multirow{2}{*}{\textbf{0.06}} \\
                                &                    & \cmark & 1.00$*$ & 1.00$*$ & 0.11$*$                  & 1.00$*$ &                                                          \\ \hline
    \multicolumn{3}{l}{MODE win percentage (2-hour)}          & \multicolumn{5}{c}{24/24}                                                                                                                                                         \\\hline
    \multicolumn{3}{l}{MODE win percentage (4-hour)}          & \multicolumn{5}{c}{14/16}                                                                                                                                                         \\ \hline
    \end{tabular}
    \begin{tablenotes}
        \item[]{$\dagger$ marks the variants that MODE outperforms significantly at p-value$<$0.05 over paired samples $t$-test. The best results are \textbf{bolded}.}
         % \item[]{}
      \end{tablenotes}

\end{table}
        
  \begin{table}[htbp]
    \centering
    \setlength\tabcolsep{1.8pt}
    \caption{Performance comparison over offline dataset.}

    \begin{tabular}{llclllll}
    \hline
    Interval                    & H            & $\mathrm{cumsum}$                & TCN                    & Seq2Seq                & Prophet                & Autoformer               & MODE                                                     \\ \hline
    \multirow{4}{*}{Interval 1} & \multirow{2}{*}{2} & \xmark & 1.65$*$ & 5.94$*$ & 1.38$*$ & 1.00$*$ & \multirow{2}{*}{\textbf{0.27}} \\
                                &                    & \cmark & 1.00$*$ & 1.00$*$ & 0.58$*$ & 1.00$*$ &                                                          \\ \cline{2-8} 
                                & \multirow{2}{*}{4} & \xmark & 1.34$*$ & 4.56$*$ & 1.46$*$ & 1.00$*$ & \multirow{2}{*}{\textbf{0.49}} \\
                                &                    & \cmark & 1.00$*$ & 1.00$*$ & 0.59$*$ & 1.00$*$ &                                                          \\ \hline
    \multirow{4}{*}{Interval 2} & \multirow{2}{*}{2} & \xmark & 0.735$*$ & 1.62$*$ & 0.65$*$ & 1.00$*$ & \multirow{2}{*}{\textbf{0.05}} \\
                                &                    & \cmark & 1.00$*$ & 1.00$*$ & 0.33$*$ & 1.00$*$ &                                                          \\ \cline{2-8} 
                                & \multirow{2}{*}{4} & \xmark & 0.704$*$ & 1.47$*$ & 0.70$*$ & 1.00$*$ & \multirow{2}{*}{\textbf{0.10}} \\
                                &                    & \cmark & 1.00$*$ & 1.00$*$ & 0.36$*$ & 1.00$*$ &                                                          \\ \hline
    \multirow{2}{*}{Interval 3} & \multirow{2}{*}{2} & \xmark & 0.66$*$ & 1.074$*$ & 0.49$*$ & 1.00$*$ & \multirow{2}{*}{\textbf{0.05}} \\
                                &                    & \cmark & 1.00$*$ & 1.00$*$ & 0.28$*$ & 1.00$*$ &                                                          \\ \hline
    \multicolumn{3}{l}{MODE win percentage (2-hour)}                         & \multicolumn{5}{c}{24/24}                                                                                                                                        \\ \hline
    \multicolumn{3}{l}{MODE win percentage (4-hour)}                         & \multicolumn{5}{c}{16/16}                                                                                                                                        \\ \hline
    \end{tabular}
    \begin{tablenotes}
        \item[]{ $\dagger$ marks the variants that MODE outperforms significantly at p-value$<$0.05 over paired samples $t$-test. The best results are \textbf{bolded}.} 

      \end{tablenotes}
\label{OfflineResult}
\end{table}   
    
In this subsection, we want to demonstrate the model performance as stated in Section \ref{MonotonicityIncreasingSection}, and answer \textbf{"Baseline Comparison"}. The synthetic and off-line experiment results are reported in Table \ref{NewSimulationData} and \ref{OfflineResult}, respectively.
    % , where the best performance model has been bolded, and the "*" marks the best model outperforming significantly at p-value $<$ 0.01 over paired samples t-test.
    We can have the following observation from Table \ref{NewSimulationData} and \ref{OfflineResult}:
    \begin{enumerate}
        \item{The MODE model can outperform the vanilla baseline models on the synthetic and offline datasets.}
        \item{With the help of the $\mathrm{cumsum}$ operation, the baseline models in $\mathrm{cumsum}$ experiments can have better performance compared to the vanilla models.}
        \item{The MODE model can still outperform most of the baseline models in $\mathrm{cumsum}$ experiments. } 
        % \item{With the help of $\mathrm{cumsum}$ operation, the Prophet baseline model can outperform the MODE model under the $\textrm{RMSE}$ evaluation metric for most of the cases. 
        %       However, under the $\textrm{MAPE}$ evaluation metric, the MODE still outperforms the Prophet model with $\mathrm{cumsum}$ trick.  }
    \end{enumerate}
\indent Based on observation 1), we can conclude that for a TSFCD problem, predicting the growth rate instead of exact values is a more suitable choice. Drawing from observation 2), we can deduce that incorporating the time difference value into the model helps stabilize the time-series, resulting in improved performance in the $\mathrm{cumsum}$ experiment. 
However, despite this improvement, observation 3) indicates that the MODE model consistently outperforms the model used in the $\mathrm{cumsum}$ experiment in most cases. These findings demonstrate the superiority of the MODE model, thus providing an answer to question \textbf{"Baseline Comparison"}.

    \subsection{Abliation Study}
    % \indent In previous subsection, we found that the MODE can outperform most of the baseline models in the cumulative dataset under simulation and offline experiments. 
    % In this subsection, we want to further investigate what  
    In this subsection, we want to take abliation study to investigate what makes the MODE model perform well and answer \textbf{"Ablation Study"}: "Does the consideration of the monotonic increasing property improve the forecast performance on cumulative data?". We conduct two extra experiments on the synthetic dataset with the consideration of the monotonic increasing from the perspective of data and model, respectively. Detailed information about the two experiments is listed as follows:
    \begin{itemize}
        \item{w/o Data Increasing (DI): We change the sample space 
    of coefficient $\zeta$ in \eqref{simulationDataset} from $[1, 10]$ to $[-10, -1]$.
    Based on this, we ablate the data monotonic increasing property.
    % On this basis, the dataset is monotonic decreasing rather than monotonic increasing, and we call this experiment without data-increasing characteristics (w/o DIC).  
    }
    \item{w/o Model Increasing (MI): We remove the $\textrm{ReLU}$ function on the right-hand side of \eqref{PredictIntegralTerm}. 
    Based on this, we ablate the model monotonic increasing property.
    % On this basis, the model cannot promise the monotonic increasing property of the predicted value, and we call this experiment without model-increasing characteristics (w/o MIC). 
    }
    \end{itemize}
All hyper-parameter settings are consist with the MODE stated in section \ref{experimentalSettingsPart}, and the experimental results are reported in Table \ref{MonotonicIncreasingExperiment}.
Based on the findings presented in Table \ref{MonotonicIncreasingExperiment}, it was observed that the MODE model achieves significantly lower MAPE values, ranging from 7.76\% to 48.41\% lower compared to the models without incorporating the DI feature and the MI property. Additionally, the MODE model demonstrates even greater improvements, with MAPE values 66.2\% to 99.01\% lower than the aforementioned models. This phenomenon underscores the importance of both the monotonic increasing property of the data and the model for enhancing prediction performance. Furthermore, in comparison to the model without DI, the model without MI exhibits inferior performance. This observation suggests that ensuring the model's adherence to the monotonic increasing property is especially crucial when dealing with cumulative data.

\begin{table}[htbp]
        \centering
        \caption{Model Performance for the Monotonic Increasing.}

    \begin{tabular}{lllll}
    \hline
    \multicolumn{2}{l}{Experiment}         & MODE                                    & w/o DI                & w/o MI                \\ \hline
    \multicolumn{2}{l}{Data monotonic Increasing}    & \cmark                   & \xmark  & \cmark  \\ \hline
     \multicolumn{2}{l}{Model monotonic Increasing}  & \cmark                   & \cmark  & \xmark  \\ \hline
    \multirow{2}{*}{Interval 2}    & 2 H   & \textbf{0.22} & 0.25$*$ & 1.97$*$ \\
                                   & 4 H   & \textbf{0.49} & 0.48$*$ & 1.29$*$ \\ \hline
    \multirow{2}{*}{Interval 3}    & 2 H   & \textbf{0.09} & 0.15$*$ & 3.80$*$ \\
                                   & 4 H   & \textbf{0.10} & 0.32$*$ & 2.28$*$ \\ \hline
    \multicolumn{1}{l}{Interval 4} & 2 H   & \textbf{0.06} & 0.13$*$ & 6.55$*$ \\ \hline
    \end{tabular}
    \begin{tablenotes}
        \item[]{$\dagger$ marks the variants that MODE outperforms significantly at p-value $<$ 0.05 over paired samples $t$-test. The best results are \textbf{bolded}.}
        \end{tablenotes}
\label{MonotonicIncreasingExperiment}
\end{table}
  
\subsection{Irregular Sampling Scenarios}
    % \afterpage{
    \begin{figure}[htbp]
        \centering
        \includegraphics[width=0.7\textwidth]{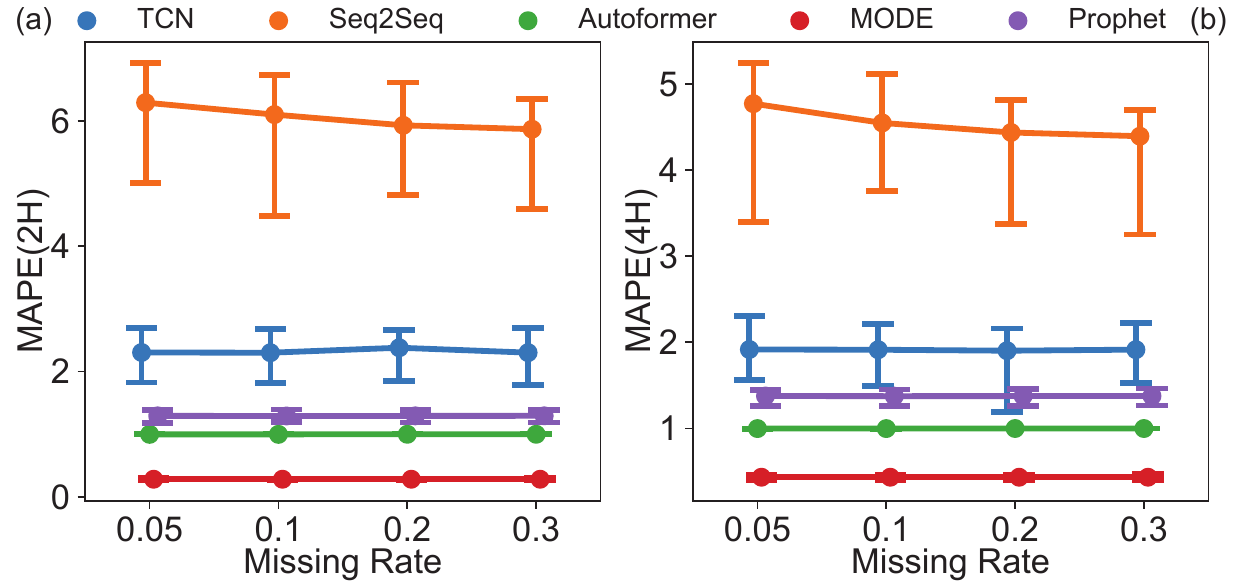}

        \caption{The irregular-sample offline experiments (a). $\textrm{MAPE}(\textrm{2H})$ at Interval 2; (b). $\textrm{MAPE}(\textrm{4H})$ at Interval 2. (We took 35\% confidence interval to highlight the main trend)}
        \label{IrregularSamplingIssues}

    \end{figure}
    
    %  }
In this subsection, we want to back up our statement about the model robustness to irregularly-sampled issues in Section \ref{IrregularityModelingStrategy} and answer \textbf{"Irregularity Sampled"}: "Is MODE robust to irregularly sampled data?". Figure \ref{IrregularSamplingIssues} presents the experimental results with irregular sampling over the offline experimental data over the missing rates in ${0.05, 0.10, 0.20}$, and $0.30$. For simplicity, we propose the evaluation metrics at Interval 2. From the Figure \ref{IrregularSamplingIssues}, we can have the following observations:
    \begin{enumerate}
        \item {The 2-hour $\text{MAPE}$ for Autoformer, Prophet, seq2seq, TCN, and MODE will fluctuate 59.05\%$\sim$62.69\%, 0.22\%$\sim$ 0.35\%,\\ 0.26\%$\sim$8.40\%, 13.49\%$\sim$ 19.84\%, and 100.19\%$\sim$100.23\%, respectively. }
        \item{The MODE model still outperforms other models under various missing data rates. The 2H $\text{MAPE}$ for the MODE is 75.56\%$\sim$93.81\% lower than other baseline models.}
    \end{enumerate}
     Observation 1) indicates that the irregularity issues will influence the model accuracy. From observation 2), we can see that even though the MODE model performance may be affected by the irregularity issue greater than other models, the MODE can still outperform other models. Hence, we back up our statement in Section~\ref{IrregularityModelingStrategy} and answer \textbf{"Irregularity Sampled"}.

\subsection{Online Experiments}
\begin{figure}[htbp]

        \centering
        \includegraphics[width=0.4\textwidth]{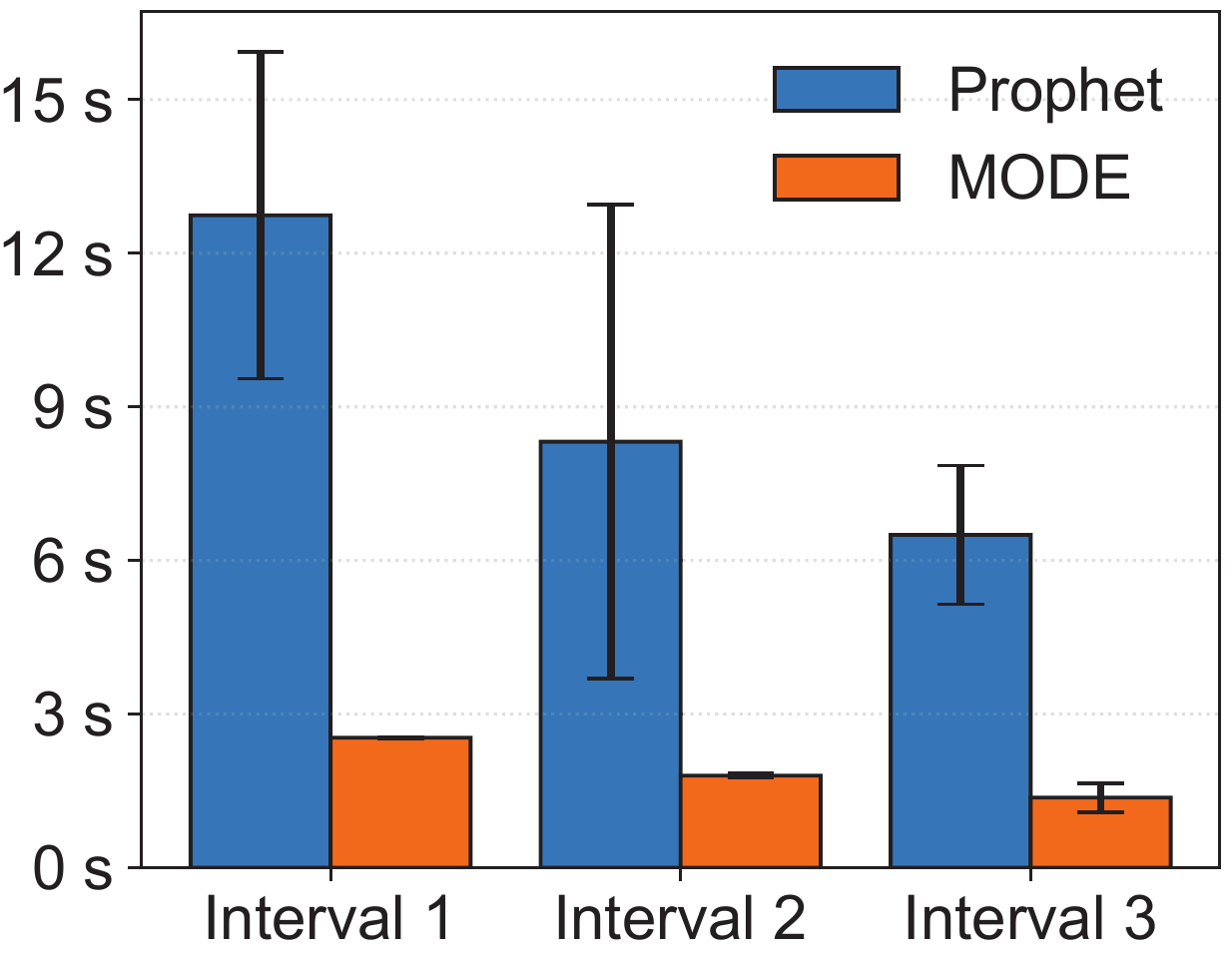}
  
        \caption{The results for online inference time}
  
        \label{InferenceTime}

        % \footnote{}
    \end{figure}

\begin{table}[htb]

\centering
\caption{Online experimental results.}

    \begin{tabular}{clllc}
    \hline
    \multirow{2}{*}{Model}   & \multicolumn{3}{c}{MAPE}                                                                                         & \multirow{2}{*}{ECP}    \\ \cline{2-4}
                             & \multicolumn{1}{c}{Interval} & \multicolumn{1}{c}{2 H}                 & \multicolumn{1}{c}{4 H}                 &                         \\ \hline
    \multirow{3}{*}{Prophet($\mathrm{cumsum}$)} & Interval 1                   & 0.11$\pm$0.09                           & 0.19$\pm$0.14                           & \multirow{3}{*}{0.13\%} \\
                             & Interval 2                   & 0.12$\pm$0.10$*$                  & 0.17$\pm$0.06$*$                  &                         \\
                             & Interval 3                   & 0.07$\pm$0.05$*$                  &                                         &                         \\ \hline
    \multirow{3}{*}{MODE}    & Interval 1                   & \textbf{0.07$\pm$0.03} & \textbf{0.12$\pm$0.06} & \multirow{3}{*}{\textbf{2.00\%}} \\
                             & Interval 2                   & \textbf{0.01$\pm$0.01} & \textbf{0.03$\pm$0.01} &                         \\
                             & Interval 3                   & \textbf{0.02$\pm$0.01} &                                         &                         \\ \hline
    \end{tabular}
    \begin{tablenotes}
        \item{$\dagger$ marks the variants that MODE outperforms significantly at p-value $<$ 0.05 over paired samples $t$-test. The best results are \textbf{bolded}.}
        \end{tablenotes}
\label{onlineResultTable}

\end{table}
Based on the previous experiments in synthetic and offline datasets, we select the Prophet($\mathrm{cumsum}$) and MODE models for the online experiment based on their $\mathrm{MAPE}$ values. In this subsection, our objective is to answer \textbf{"Online Performance"} by validating the prediction accuracy and demonstrating the business value of the MODE model. Both models were deployed on the organizers' internal Python platform. To ensure fairness, each model served the activity for six days. The $\textrm{MAPE}$ values and the ECP values are reported in Table \ref{onlineResultTable}. Additionally, the inference time of the two models is plotted in Figure \ref{InferenceTime}.
    We can obtain following observations from Table \ref{onlineResultTable} and Figure \ref{InferenceTime}:
    \begin{enumerate}
        \item {The MODE model in the online environment has a lower MAPE than the Prophet model. 
        Take the 2-hour MAPE as an example the evaluation metric values of MODE are 8.6 $\sim$ 78.44 \%, 60.36 $\sim$ 95.53 \%, and 27.4 $\sim$ 93.42 \% lower than those of the Prophet model, respectively.}
        \item{The MODE model's ECP is two orders of magnitude higher than that of the prophet model. }
        %\item{The $\text{MAPE}$ standard deviation of the MODE are 62.73\% $\sim$ 75.51\%, 56.73\% $\sim$ 81.77\%, 11.89\% $\sim$ 85.30\% lower than those of the Prophet model in for 2-hour-horizon, 4-hour-horizon, and final point, respectively.}
        \item{The inference time of the MODE in Intervals 1, 2, and 3 are 80.06\%, 78.51\%, and  78.97\% lower. 
        Specifically, the inference time of the MODE is less than 5 seconds.
        }
    \end{enumerate}
    
Observation 1) reflects the correctness of our growth rate modeling strategy.  
Observation 2) proves that when we adopt the same adjustment strategy to allocate the virtual card, 
the MODE model can promote a higher total collected headcount compared to the prophet model. This phenomenon reflects the business value of the MODE model. Observation 3) indicates that the MODE inference speed is far faster than the Prophet model. The reason for this phenomenon is that the prophet will tune the hyper-parameters with its inner optimization framework when it obtains the request, while the MODE merely needs to solve the ODE IVP.
In all, the performance of the online experiments answers \textbf{"Online Performance"}.

    % As such, we can augment the ODE as follows:
    % \begin{equation}
    %   \frac{d\hat{z}(t)}{dt} =[f_{\theta}(y(t), t), -{\frac{\partial{f_{\theta}}}{\partial{y}}}^\top a(t), -{\frac{\partial{f}_{\theta}}{\partial{\theta}}}^\top a(t)],
    %   \label{augSystem}
    % \end{equation}
    % where the first element in the vector $\hat{z}(t)$ is the original NODE model, the second term is the ODE for the adjoint state, then the last term is the ODE for the gradient with respect to parameter $\theta$.\\

    % which will result in the memory storage problem in the model training. To this end, the adjoint sensitivity method in the NODE model is 
    %  possibly making our computation graph too big to hold in memory for backprop.
    % Unfortunately, the forward solve might have taken a huge number of steps depending on the system dynamics, possibly making our computation graph too big to hold in memory for backprop.
    % 
    % , and embed the time difference information 
    
    % prediction use the cumulative summation operator on the non-negative predicted value realized via activated function like . On this basis, 

    % Previous works on time-series forecasting have realized great success in different fields. Intuitively, it is natural to model such cumulative data with conventional time-series forecasting methods such as recurrent neural networks (RNNs), temporal convolution networks (TCNs), Transformers, and attention models. However, these methods fail to capture the following properties of cumulative data, limiting their performance in industrial applications: 

\section{Conclusions}\label{ConclusionSection}
\indent In this paper, we addressed the TSFCD via MODE model within the NODE framework.
We first analyzed our task from the monotonic property. Based on the analysis, we found that it is easy to guarantee the monotonic increasing property of the predicted values via modeling its growth rate rather than its exact value. Thus, we forced the growth rate term in differential expression to be positive and proposed our MODE model within the NODE framework. Throughout converting TSFCD into an ODE IVP, we addressed the irregularity issue naturally thanks to the numerical methods for ODEs. To validate the effectiveness of our proposed method, we conducted extensive experiments on the simulation, offline, and online environments. \\
\indent There are two remaining directions for future work that can be explored. Firstly, the current approach relies mainly on the initial value alone when inferring future predicted values, disregarding the sequence of initial values (a finite-size window that captures past values). This heavy dependence on the initial value may expose the model training process to data fluctuation issues. Therefore, an interesting and promising direction would be to investigate the integration of neural controlled differential equations \citep{kidger2020neural} with the prediction model. 
% This combination could potentially address the limitations of relying solely on the initial value.
Secondly, the adjustment of the coupon allocation strategy can be framed as a model predictive control (MPC) problem. To enhance decision-making, it would be beneficial to incorporate relevant indices from practical downstream scenarios into the MPC objective function. By doing so, we can make more informed decisions. Building upon this foundation, it would be captivating to explore a more seamless integration of decision algorithms and forecasting models, aiming for a tighter and more effective combination.

\newpage
\bibliographystyle{unsrtnat}
\bibliography{reference_new}
\end{document}